\documentclass[graybox]{svmult}

\usepackage{type1cm}
\usepackage{makeidx}
\usepackage{graphicx}
\usepackage{multicol}
\usepackage[bottom]{footmisc}
\usepackage{newtxtext}
\usepackage[varvw]{newtxmath}

\usepackage[caption=false]{subfig}
\usepackage{amsmath}
\usepackage{amsfonts}

\usepackage{hyperref}
\usepackage{cleveref}

\makeindex

\begin{document}

\title*{Enabling Motion-Induced WiFi-CSI Sensing in Multi-Robot Systems: A Simulation Framework}

\titlerunning{Enabling Motion-Induced WiFi-CSI Sensing in Multi-Robot Systems}

\author{Brendan Dijkstra\inst{1}, Ninad Jadhav\inst{2}, Maurice Stoffijn\inst{1}, Stephanie Gil\inst{2}, and Bahar Haghighat\inst{1}}

\institute{\inst{1}University of Groningen, 9747 AG Groningen, The Netherlands \and \\ \inst{2}Harvard University, Cambridge, MA 02138, USA.}%

\authorrunning{Dijkstra et al.}

\maketitle

\abstract{%
WiFi Channel State Information (CSI) enables fine-grained sensing of spatial relationships between wireless devices. The WiFi Sensor for Robotics (WSR) toolbox demonstrated that WiFi-CSI measurements collected during robot motion, combined with odometry, can be used to synthesize virtual antenna arrays, enabling motion-induced bearing estimation in multi-robot experiments. However, deploying the WSR approach has remained challenging due to its limited integration with standard robotics tools and reliance on deprecated WiFi-CSI acquisition hardware. This work presents an open-source simulation framework that integrates the WSR sensing pipeline within the robotic simulator Gazebo. The framework generates synthetic WiFi-CSI measurements with realistic physical signal properties and leverages realistic robot motion required for motion-induced sensing and downstream localization. It further supports processing both simulated and real WiFi-CSI data for direct comparison and validation. Our aim is to make WSR-style sensing accessible as a modular sensing modality for robotic research. 
We validate our framework by comparing simulated and real-world experiments using TurtleBot3 robots, and demonstrate its use for bearing estimation and robot localization. This work lowers the barrier to entry for WiFi-CSI-based sensing in multi-robot systems and provides a foundation for integrating WiFi-CSI sensing with multi-modal perception for localization and coordination. All code and experimental details are publicly available at \url{https://github.com/baharhaghighat/DARS_2026_Dijkstra_et_al}
}

\def\arraystretch{1.15}

%
%

\section{Introduction}
\label{sec:intro}

Multi-robot systems rely on accurate relative sensing to support localization, coordination, and autonomous operation in complex environments \cite{Cadena2016PastAge, Burgard2000CollaborativeExploration, Stump2011Multi-robotProblem, Chen2015AdaptiveVision, Fenwick2002CooperativeLocalization, Jadhav2022ToolboxRobotics}. In indoor and GPS-denied settings, obtaining such information remains challenging. Existing solutions based on vision systems, Ultra-Wideband (UWB) radios, or Bluetooth beacons can provide accurate relative sensing information, but often require dedicated hardware, careful deployment, or favorable environmental conditions \cite{Prorok2013AccurateCollaboration, Martinelli2020CooperativeCases}. At the same time, the growing deployment of multi-robot systems in domains such as warehouse automation and industrial environments continues to drive the need for scalable and easily deployable indoor sensing solutions \cite{Arents2022SmartManufacturing, Queralta2020CollaborativeVision, Tantawi2019AdvancesCollaboration, Chakraborti2020FromForum}.

WiFi signals offer an attractive alternative sensing modality. WiFi infrastructure is already widely deployed in most indoor environments, offering a cost-effective and scalable foundation. Unlike GPS, WiFi signals can penetrate walls and operate in both Line-of-Sight (LOS) and non-line-of-sight (NLOS) conditions, making them suitable for complex indoor settings where signal occlusion is common.

WiFi-based localization has traditionally relied on Received Signal Strength Indicator (RSSI), which estimates distance based on signal attenuation \cite{Zhu2013RSSI-basedLocalization, Coppola2018On-boardTeams, Amoolya2019Wi-FiEnvironment}. However, RSSI is susceptible to noise and multipath effects, common in indoor environments \cite{Jadhav2022ToolboxRobotics, Zafari2019ATechnologies, Yan2013Co-OptimizationEnvironments}.
More recent work leverages Channel State Information (CSI), which captures fine-grained amplitude and phase data across multiple subcarriers.

CSI has enabled a broad range of sensing applications, including imaging and tracking through obstacles \cite{Adib2015CapturingWall, Depatla2017RoboticImaging, Adib2013SeeWi-Fi}, human activity and gesture recognition \cite{Jin2018WiSh:RFIDs, CominelliExposingLimitations, Wang2019AInformation}, and material classification \cite{Zhang2019OnSensing, Ha2020FoodRFIDs, Dhekne2018TrackIO:Inside-Out}. Despite the wide applicability, existing methods often overlook the challenges and opportunities posed by mobility of transceiver nodes, limiting their integration with mobile robotic platforms \cite{hunter_wiros_2023}.

The WiFi Sensor for Robotics (WSR) toolbox is a key exception. WSR demonstrated that WiFi-CSI measurements collected during robot motion, together with robot odometry, can synthesize a virtual antenna array and produce Angle-of-Arrival (AoA) profiles between robots \cite{Jadhav2022ToolboxRobotics, Jadhav2022ARobotics}. This motion-induced sensing approach allows robots to estimate relative bearing without physical antenna arrays and can operate with arbitrary 2D or 3D motion, including cases where both transmitting and receiving robots move during data acquisition \cite{Jadhav2022ToolboxRobotics, Jadhav2022ARobotics}. WSR therefore provides a promising sensing primitive for multi-robot systems, where robots can exploit their natural motion to obtain relative direction information from communication signals. Despite this promise, WSR-style sensing remains difficult to deploy and study broadly. The original WSR implementation relies on Intel 5300 WiFi hardware and modified low-level drivers for CSI acquisition \cite{Halperin2011ToolRelease, Jadhav2022ARobotics}. This hardware stack is increasingly deprecated and hard to source, and the full setup requires substantial integration effort across wireless drivers and robot platforms. These constraints limit reproducibility and make it difficult to investigate how WiFi-CSI sensing interacts with robot trajectory design, localization pipelines, and other sensing modalities.

Simulation is essential for studying such interactions. While MATLAB’s WLAN Toolbox provides wireless channel modeling tools, it is intended for static scenarios and does not model the continuous robot motion or environmental dynamics. Conversely, platforms like NVIDIA Aerial Omniverse offer highly detailed physical layer simulation, but they are tailored for large-scale 5G/6G networks and require significant computational resources. As a result, there is currently no accessible simulation framework that reproduces the WSR sensing pipeline while allowing multi-robot operation, CSI generation, bearing estimation, and downstream localization.

\begin{figure}[t]
    \centering
    \includegraphics[width = 0.95\textwidth]{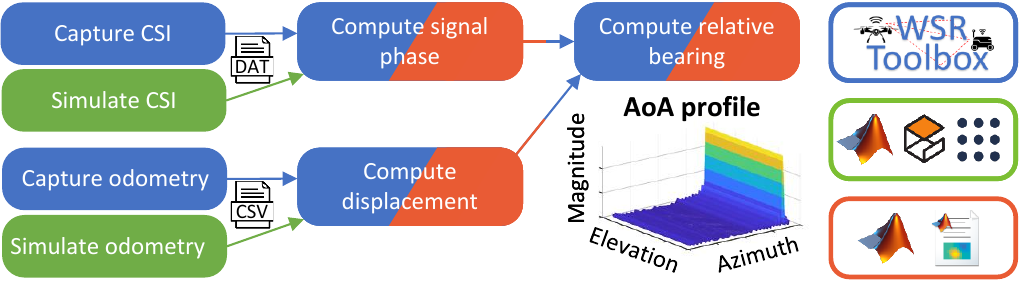}
    \caption{Overview of the functionality of our simulation framework (green and orange) in parallel with that of the WSR toolbox (blue). The framework blocks in green involve implementations in MATLAB, Gazebo, and ROS and those in orange involve MATLAB. The WSR toolbox blocks in blue involve implementations in C++ and Python from \cite{Jadhav2022ToolboxRobotics}. Our simulation framework operates fully independently from the WSR toolbox software package as we directly implement its methods within MATLAB. Three steps are included in our framework: (i) simulating raw CSI and robot odometry data, (ii) computing phase using CSI data, and (iii) running Bartlett's estimator and Levenberg–Marquardt algorithm to estimate robots' relative bearing and position.}
    \label{img:tools_overview}
\end{figure}

This paper addresses this gap by presenting an open-source simulation framework
for motion-induced WiFi-CSI sensing in multi-robot systems. The framework
integrates MATLAB-based CSI signal processing with Gazebo-based robot
simulation and reproduces the WSR sensing pipeline, including CSI and odometry
generation, phase computation, virtual array construction, AoA estimation using
Bartlett’s estimator, and downstream localization (see Figure~\ref{img:tools_overview}). The contributions are threefold:
\begin{itemize}
    \item An open-source simulation framework that reproduces the WSR sensing
    pipeline for motion-induced WiFi-CSI sensing in multi-robot systems.
    \item Integration of WiFi-CSI sensing with Gazebo-based robot simulation,
    enabling joint evaluation of sensing, motion, and multi-robot configurations.
    \item Experimental validation using both simulated data and real-world
    measurements.
\end{itemize}

%
%

\section{The WSR Toolbox}

The WSR Toolbox provides an open-source solution for real-time WiFi-CSI-based localization, rendezvous, and mapping on mobile robotic platforms \cite{Jadhav2022ARobotics, Jadhav2022ToolboxRobotics}. It enables bearing estimation by transforming arbitrary 2D or 3D robot trajectories into a virtual antenna array, making it well-suited for mobile and multi-robot applications.

The WSR hardware setup is built around an Intel WL5300 Network Interface Card (NIC), which is operated using custom drivers developed by the authors using open source firmware to extract CSI data \cite{Halperin2011ToolRelease}. This NIC is installed on a single-board computer (SBC) that can be mounted on a variety of ground or aerial robots \cite{Jadhav2022ARobotics, Jadhav2022ToolboxRobotics}.

\begin{figure}[t]
    \centering
    \includegraphics[width=0.8\textwidth]{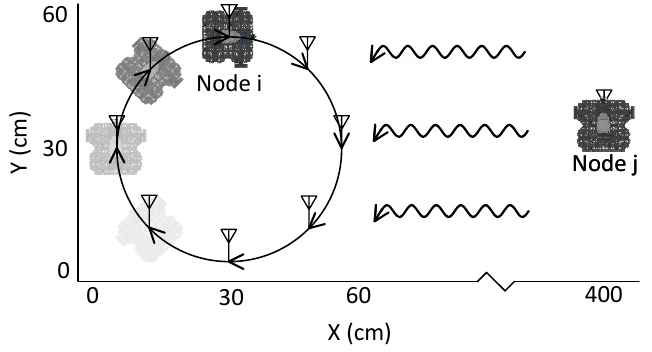}
    \caption{Creating a virtual antenna array by moving a single physical antenna element hosted on a mobile robot. This illustration shows (i) a circular virtual antenna array with a 30~cm radius and (ii) a transmitting node placed 4~m away to satisfy the far-field assumption considered in \cite{Jadhav2022ARobotics}.}
    \label{fig:antenna-array}
\end{figure}

To create the virtual antenna array, the toolbox uses a Synthetic Aperture Radar (SAR) approach: as illustrated in \Cref{fig:antenna-array}, a single antenna is moved along a trajectory while collecting CSI measurements. Unlike a physical antenna array that samples a single transmission across multiple spatially separated elements, the virtual array requires multiple transmissions, one for each sampling location as the robot moves, and utilizes robot odometry information to reconstruct the spatial geometry of the sampled positions \cite{Jadhav2022ARobotics}.

The toolbox handles the packet exchange between the nodes to obtain channel properties via both forward and backward transmissions, capturing CSI for 30 of the 48 WiFi subcarriers available under the 802.11n standard \cite{Gil2015AdaptiveStrength}. The central subcarrier is interpolated out of these measurements after which the signal phase is extracted \cite{Halperin2011ToolRelease,Sen2012YouInformation,Jadhav2022ToolboxRobotics}.
By extracting position information from the odometry, an Angle-of-Arrival (AoA) profile is constructed as shown in \Cref{subsec:bearing}. The AoA profile maps incoming signal power as a function of direction across the incoming directions on a given resolution. In two-dimensional setups, the profile includes azimuth angles; in three-dimensional setups, it includes elevation as well. The direction with the highest signal strength, referred to as AoA\textsubscript{max}, can be selected as the estimated bearing between the nodes or multiple observations can be used, for example, for target tracking purposes. A likelihood score, based on the magnitude of the AoA profile at AoA\textsubscript{max}, can be used to quantify the confidence in the estimate. The WSR toolbox thus delivers a complete processing pipeline, from packet transmission coordination and CSI data capturing to AoA profile generation and bearing estimation on real hardware \cite{Jadhav2022ARobotics}.
Despite its versatility, deploying the WSR toolbox has been limited due to the reliance on obsolete hardware, the complexity of the corresponding software which has been end-of-life since 2018, and the large integration effort required to establish a working multi-robot setup.

%
%

\section{Proposed simulation framework}
\label{sec:sim_framework}
Our proposed simulation framework replicates the operation of the WSR Toolbox and comprises three main stages: (i) capturing raw CSI and robot odometry, (ii) computing signal phase and virtual array geometry, and (iii) relative bearing estimation via Bartlett’s estimator (see \Cref{img:tools_overview}). We additionally present an application study on absolute position estimation using the Levenberg–Marquardt algorithm which can easily be extended to different kinds of multi-robot research.

The framework runs on Ubuntu and leverages tight integration between Gazebo and ROS to simulate robotic behavior and wireless signal propagation in real time. It supports two compatibility layers: initially developed and validated on Ubuntu 20.04, MATLAB 2022a, ROS1 with Gazebo Classic, and later upgraded for compatibility and speed to Ubuntu 24.04, MATLAB 2025b, ROS2 with GZ Gazebo. The latter is used for the results in this paper.

\begin{figure}[t]
\centering
\subfloat[Real robot with onboard WiFi module.\label{fig:robot-real}]{%
  \includegraphics[width=0.64\textwidth]{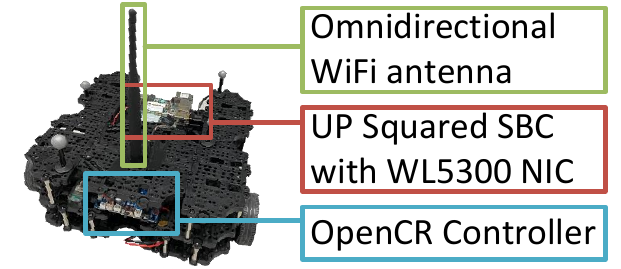}%
}\hfill
\subfloat[Simulated robot.\label{fig:robot-sim}]{%
  \includegraphics[width=0.29\textwidth]{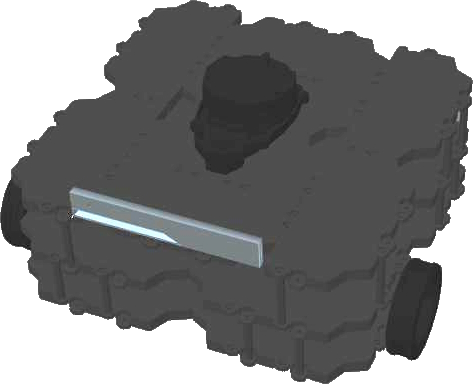}%
}
\caption{We use TurtleBot3 Waffle robots in our simulation and real setups to collect data. (a) Real robot with onboard UP Squared SBC, Intel WL5300 NIC, and WiFi antenna. (b) Simulated robot with the WiFi antenna at the model's center.}
\label{fig:robot}
\end{figure}

MATLAB interfaces with Gazebo using the ROS Toolbox. It handles robot spawning, trajectory control, data collection, and all CSI processing tasks, including bearing estimation via Bartlett’s estimator followed by the minimization solver for localization. Simulated data is saved in the same format as real datasets, enabling direct interchangeability between simulation and experimental pipelines for validation purposes. Robots in simulation use the same ROS interfaces as physical robots and simulate drift accurate to the specific robotic platform, ensuring high fidelity and consistency.
We use the TurtleBot3 Waffle platform for both simulated and real-world setups, as shown in \Cref{fig:robot}.

\subsection{Signal Model}
\label{subsec:signal_model}

Under the ideal channel assumption between nodes \( i \) and \( j \), the central carrier is modeled as a monochromatic electromagnetic wave, where the signal amplitude attenuates with distance and the phase encodes spatial separation:

\begin{equation}
    h_{ij}(t) = \frac{1}{d_{ij}(t)}e^{-2\pi i\frac{d_{ij}(t)}{\lambda}}
    \label{eq:ideal-channel-signal}
\end{equation}

Here, \( d_{ij}(t) \) is the Euclidean distance between nodes at time \( t \), and \( \lambda \) is the signal wavelength. As distance is symmetric (\( d_{ij} = d_{ji} \)), we use a single subscript notation.

\subsection{Simulating CSI Data}
\label{subsec:csi_sim}

We simulate CSI data based on real-world insights to ensure high fidelity in our framework. The primary focus is on modeling the CSI phase and robot odometry, which are crucial components for the AoA profile. The simulated CSI incorporates four elements:

\subsubsection*{Carrier Frequency Offset (CFO)}
\label{subsubsec:cfo}
CFO arises from slight mismatches in the transmitter and receiver's oscillator frequencies. The modeled CSI from node \( i \) to node \( j \), denoted as \( \hat{h}_{ij}(t) \), incorporates CFO and additive noise \(N\) as follows:

\begin{equation}
\begin{split}
    \hat{h}_{ij}(t) &= \frac{1}{d_{ij}(t)} e^{-2\pi i ( \frac{d_{ij}(t)}{\lambda} + \textit{CFO}_{ij}(t) + N)} \\
    \hat{h}_{ji}(t) &= \frac{1}{d_{ij}(t)} e^{-2\pi i ( \frac{d_{ij}(t)}{\lambda} - \textit{CFO}_{ij}(t) + N)}
\end{split}
\label{eq:cfo-channel-signal}
\end{equation}

Assuming channel reciprocity, CFO-induced phase offsets appear with equal magnitude but opposite signs for \( \hat{h}_{ij} \) and \( \hat{h}_{ji} \) and can be removed by multiplying the signals \cite{Jadhav2022ToolboxRobotics,rahul_jmb_2012}:

\begin{equation}
    h_{ij}^2(t) = h_{ji}^2(t) = \hat{h}_{ij}(t) \times \hat{h}_{ji}(t)
    \label{eq:csi-multiplied}
\end{equation}

Each virtual antenna uses a unique fixed CFO, replicating a real hardware offset.

\subsubsection{Additive Phase Noise}
\label{subsubsec:additive_noise}
After CFO cancellation, residual phase noise persists due to environmental factors like antenna variance, multipath effects, and signal interference. We approximate this residual as Additive White Gaussian Noise (AWGN), using MATLAB's standard AWGN channel model. The noise variance $\sigma^2$ is determined by the average signal power $P_{\text{signal}}$ and the specified Signal-to-Noise Ratio (SNR).

\begin{equation}
\text{SNR} = 10 \log_{10} \left( \frac{P_{\text{signal}}}{\sigma^2} \right) 
\end{equation}
Consequently, the noise variance is calculated as follows and can be set to different SNR levels for each connected node pair:
\begin{equation}
\sigma^2 = \frac{P_{\text{signal}}}{10^{\text{SNR}/10}} 
\end{equation}

\subsubsection{Odometry Drift}
\label{subsubsec:odom_drift}
By simulating robotic hardware in the 3D robotic simulator Gazebo, we include robotic drift corresponding to the specific robotic platform. The odometry drift is included by using the true position of the robot for simulation of the WiFi CSI signal data and using the odometry for calculating the bearing, matching real conditions.

\subsubsection{Randomization in Repeated Experiments}
\label{subsubsec:real_deviations}
The starting positions of the moving robots are randomized in a range of $\pm20$ cm in any direction. This realistically replicates small offsets that occur when manually repeating experiments with real hardware.

\subsection{Estimating AoA Profile}
\label{subsec:bearing}

Under the far-field assumption, we treat incoming wavefronts as planar over the robot displacement.
The steering vector \( \mathbf{a}_{i}(\theta, \phi)(t) \) encodes the expected phase shifts for each possible discretized arrival direction of \( (\theta, \phi) \), defined as \cite{Jadhav2022ToolboxRobotics}:

\begin{equation}
    \mathbf{a}_{i}(\theta, \phi)(t) =
    {e}^{-2\pi i \frac{d_{i}(t)}{\lambda}\sin\theta\sin\xi_i(t)\cos(\phi-\varphi_i(t))+\cos\xi_i(t)\cos\theta}
    \label{eq:steering_vector}
\end{equation}

Here, \( d_i(t) \) is the displacement of robot \( i \) between \( t_0 \) and \( t \), while \( \varphi_i(t) \) and \( \xi_i(t) \) denote spherical coordinate angles for the robot's heading. The AoA profile is computed by projecting the CSI signal onto all steering directions:

\begin{equation}
    F_{ij}(\theta, \phi) = \left| \sum_{t=t_0}^{t_{end}} h_{ij}^2(t) \times \mathbf{a}_{i}^2(\theta, \phi)(t) \right|^2
    \label{eq:bartlett}
\end{equation}

Equation \eqref{eq:bartlett} is Bartlett's estimator \cite{Krim1996TwoApproach}. The AoA profile reveals peaks at directions where the received signal energy is strongest, indicating the direction of the transmitting node. We discretize azimuth \( \theta \in [-180^\circ, 180^\circ] \) and elevation \( \phi \in [-90^\circ, 90^\circ] \) at \(1^\circ\) resolution, following \cite{Jadhav2022ARobotics, Jadhav2022ToolboxRobotics}.

\subsection{Application Study: Localization}

We estimate the mobile receiver's position $\mathbf{p} \in \mathbb{R}^2$ from a set of static and mobile transmitters, whose 2D projected positions are defined as $\mathcal{Q} = \{\mathbf{q}_i \mid i=1, \dots, K\}$, where $K$ is the total number of transmitters. This is solved via nonlinear least-squares by minimizing the squared perpendicular distances from $\mathbf{p}$ to the $AoA_{max}$ bearing lines, each defined by $\mathbf{q}_i$ and a direction vector $\mathbf{n}_i = [\cos\theta_i, \sin\theta_i]^\top$ derived from bearing $\theta_i$.

The perpendicular distance from candidate position \(\mathbf{p}\) to the bearing line is given in \Cref{eq:distance} and the optimization problem is formulated in \Cref{eq:optimization}.
\begin{equation}
d_i(\mathbf{p}) = \left\| (\mathbf{I} - \mathbf{n}_i \mathbf{n}_i^\top) (\mathbf{p} - \mathbf{q}_i) \right\|_2
\label{eq:distance}
\end{equation}

\begin{equation}
\hat{\mathbf{p}} = \arg \min_{\mathbf{p}} \sum\nolimits_{i=1}^{K} \left[ d_i(\mathbf{p}) \right]^2
\label{eq:optimization}
\end{equation} 

This nonlinear least-squares problem is solved using the Levenberg--Marquardt algorithm, initialized at the centroid of active transmitters \cite{watson_levenberg-marquardt_1978, levenberg_method_1944, marquardt_algorithm_1963}:
\begin{equation}
\mathbf{p}_0 = \frac{1}{K} \sum\nolimits_{i=1}^{K} \mathbf{q}_i
\label{eq:solver}
\end{equation}

%
%

\section{Experiments \& Results}
\label{sec:experiments}

We evaluate our simulation framework on both AoA and localization performance across three scenarios of increasing complexity, designed to test accuracy, scalability, and generalization. Each scenario is implemented using a set of mobile and static TurtleBot3 Waffle Pi robots performing 100 Monte Carlo simulations. The first scenario in \Cref{subsec:benchmark} establishes a baseline with 2D circular motion, allowing direct comparison with real-world results from an identical setup presented in \cite{Jadhav2022ToolboxRobotics}. The second scenario in \Cref{subsec:expanded} extends the setup to a 3D environment with elevated nodes and multiple mobile agents by dynamically creating platforms in Gazebo according to the 3D trajectories configured in MATLAB. The third scenario in \Cref{subsec:generalization} introduces random walk trajectories to evaluate performance under unstructured motion.

\subsection{Benchmark Experiment: 2D Circular Motion}
\label{subsec:validation_experiments}
\label{subsec:benchmark}

\begin{figure}[t]
    \centering
    \includegraphics[width=1\linewidth]{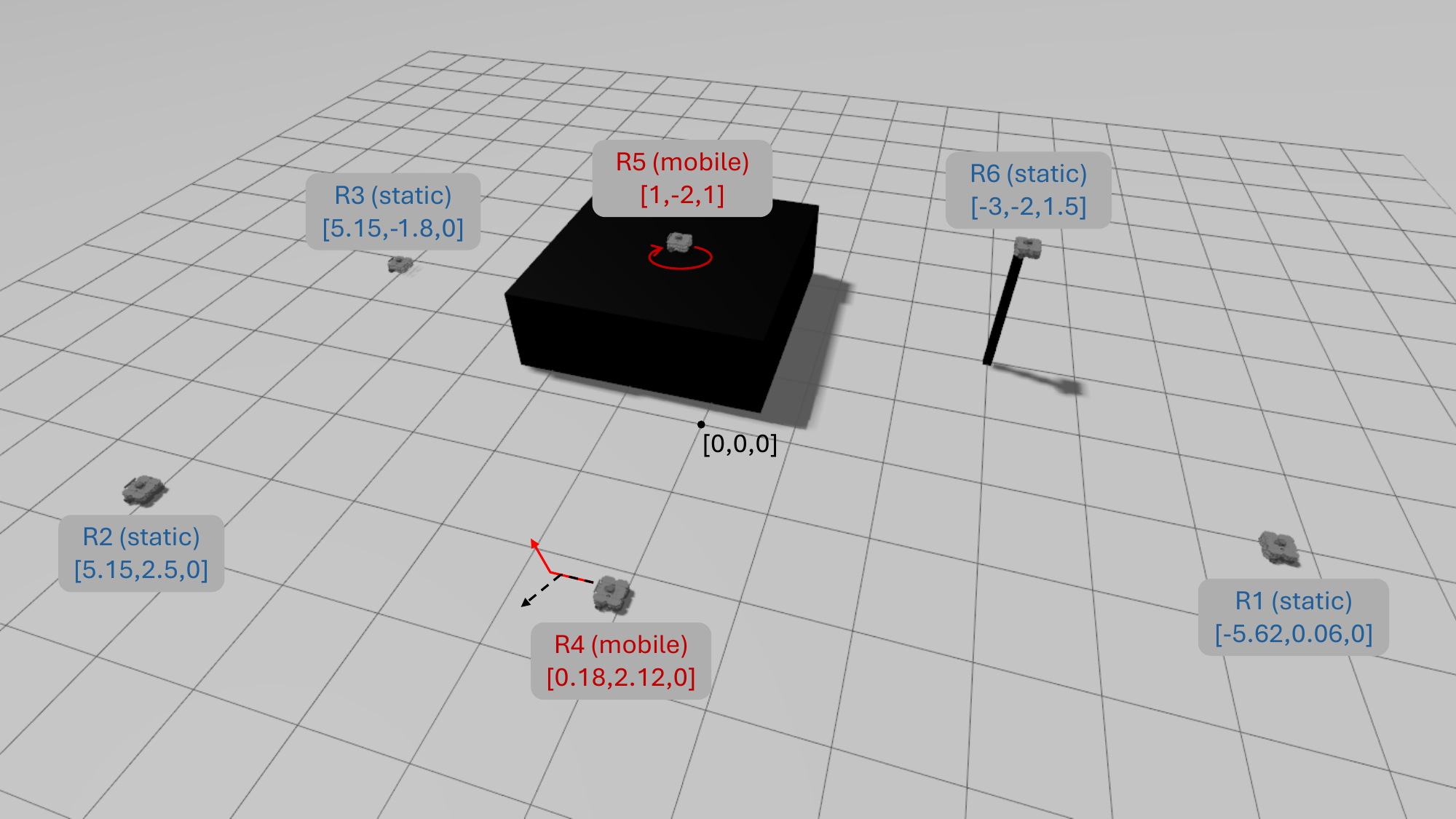}
    \caption{3D view of the simulated experiment setup in Gazebo with an indication of the trajectories. The benchmark scenario matching the setup described in \cite{Jadhav2022ToolboxRobotics} only includes R1 to R4, while R5 and R6 are added in the experiments that follow. R4 and R5 are mobile while R1, R2, R3, and R6 are static beacons.}
     \label{fig:experiment-setup-gazebo-3d}
\end{figure}

This baseline scenario replicates the scenario studied in \cite{Jadhav2022ToolboxRobotics} and involves four robots: three static (R1--R3) and one mobile (R4), all placed at the ground level. The \((x,y)\) coordinates of the four robots are shown in \Cref{fig:experiment-setup-gazebo-3d}. R4 follows a circular trajectory with a 20 cm diameter using a forward velocity of 0.04~m/s and a rotational velocity of 0.2~rad/s. This trajectory exceeds the minimum displacement required for successful bearing estimation, i.e., twice the wavelength at 5~GHz, approximately 12~cm \cite{orfanidis_electromagnetic_2016,Jadhav2022ARobotics}.
We ensure spatial resolution finer than \(\lambda/2 = 27\) mm by sampling CSI at 100 Hz for the given speed and wavelength. The experiment is carried out over 3 seconds of simulated time, which is longer in real time as the speed is limited by the performance of the hardware. All node pairs experience a low noise level with an SNR of 30, which should approximate the LOS scenario presented in \cite{Jadhav2022ToolboxRobotics}.

\subsubsection*{Benchmark Scenario Results}

We benchmark the simulation framework by comparing a simulated circular trajectory with real-world data and assess the impact of different noise components.  

First, we evaluate using the true robot position for Bartlett computation instead of the robot odometry. We obtain a median error of $0.05$ m and a standard deviation of $0.02$ m which shows that the system is able to localize the target to high accuracy under ideal conditions. 

Next, we evaluate the same runs using odometry captured by the robot which yields a median positioning error of $0.42$ m, showing that the robotic drift is a significant factor of error.
The median is relatively close to the $0.50$ m reported in real experiments reported by \cite{Jadhav2022ToolboxRobotics}. 
However, the standard deviation is quite low in the simulation at $0.17$ m because the main component of error, the odometry drift, is relatively deterministic given a fixed trajectory.

\subsection{Expanded Experiment: Elevated and Multiple Mobile Agents}
\label{subsec:expanded}

This scenario expands the setup into a three-dimensional multi-robot experiment by introducing two additional robots at different elevations. The new nodes include one mobile (R5) and one static (R6), as depicted in \Cref{fig:experiment-setup-gazebo-3d}. The mobile nodes will perform the same trajectory as the benchmark scenario and estimate their position relative to the four static nodes. We set the SNR to 3 for robot pairs [1,4], [1,5], [4,6], and [5,6], mimicking a NLOS scenario where R1 and R6 are in different rooms or otherwise obstructed from direct view.

This configuration increases spatial diversity and decentralization, allowing us to evaluate system performance under multiple simultaneous localization tasks and more realistic multi-robot deployments. The spatial diversity creates ambiguity in the bearing estimation over the elevation angle, as the trajectory itself is confined to a 2D plane and does not contain an elevated 3D component to inform the estimation. This effect is reported in \cite{Jadhav2022ToolboxRobotics} as trajectory informativeness.

\subsubsection*{Expanded Scenario Results}
\begin{figure}[t]
\centering
\subfloat[AoA profile for R4 to R3 using odometry\label{fig:aoa_r4_r2}]{%
    \includegraphics[width=0.50\textwidth]{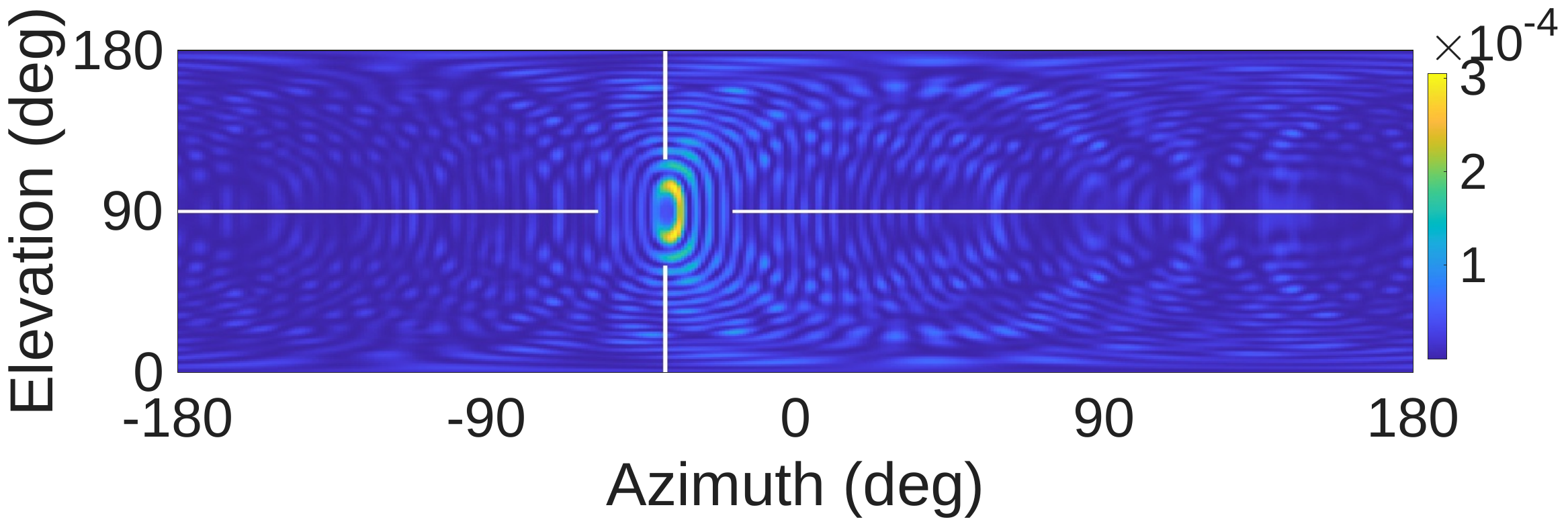}%
}\hfill
\subfloat[AoA profile for R4 to R6 using odometry\label{fig:aoa_r5_r3}]{%
    \includegraphics[width=0.50\textwidth]{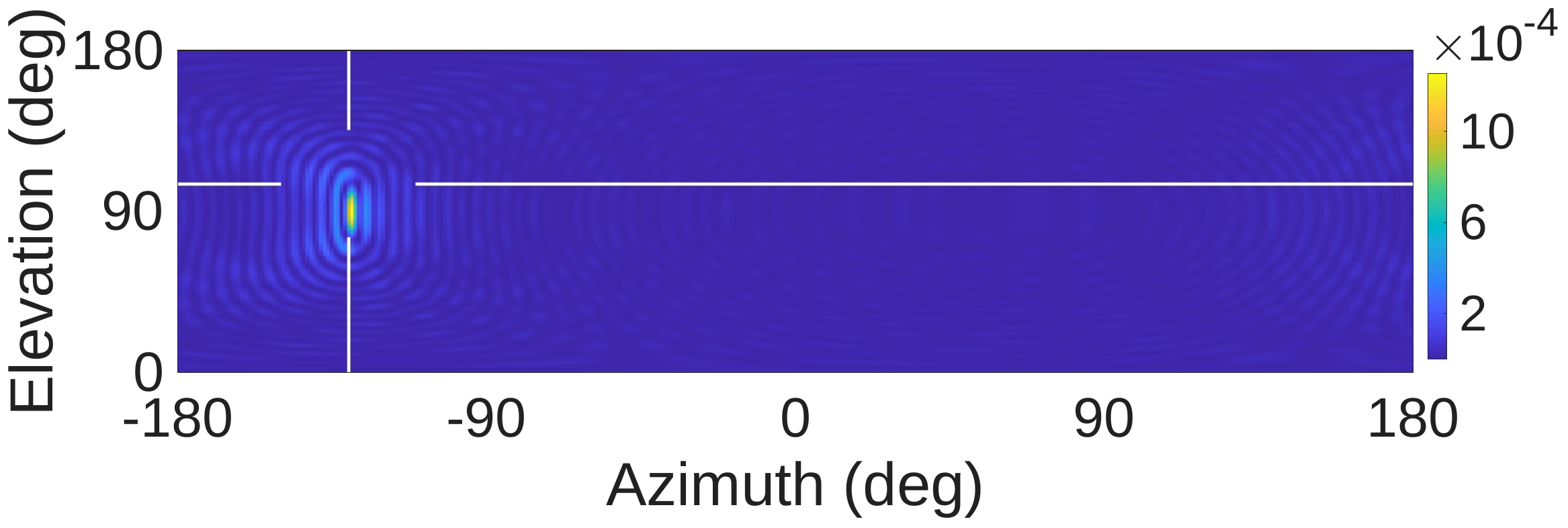}%
}

\vspace{0.5em}

\subfloat[AoA profile for R4 to R3 using true pose\label{fig:aoa_r4_r5}]{%
    \includegraphics[width=0.50\textwidth]{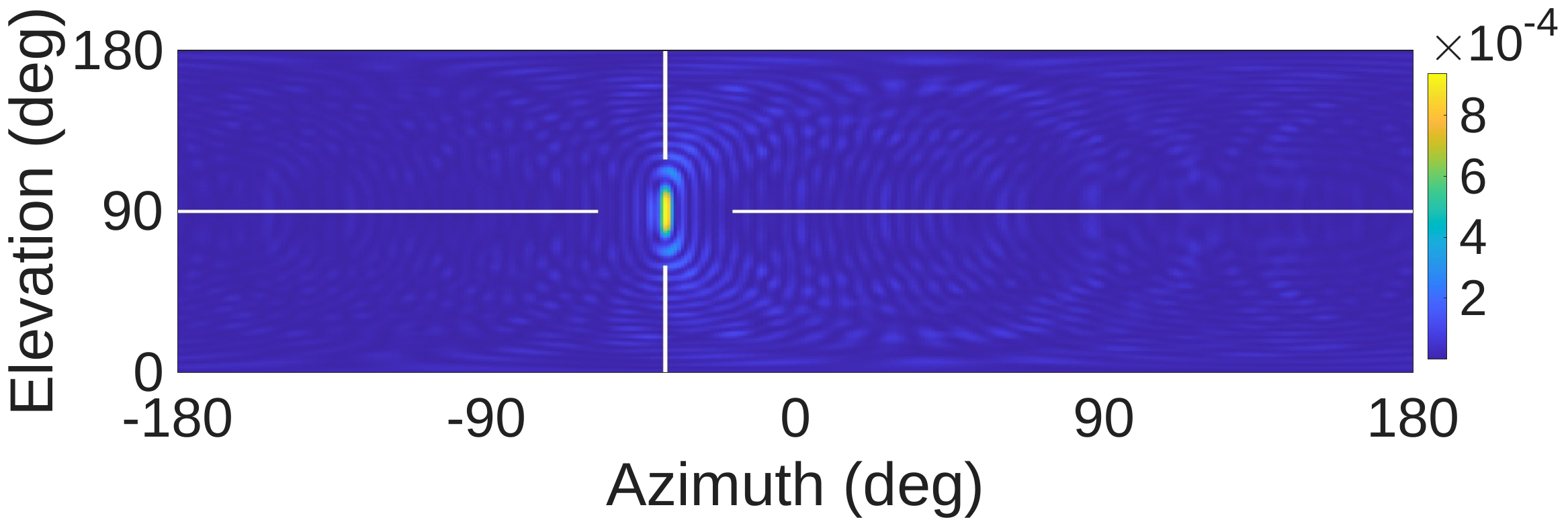}%
}\hfill
\subfloat[AoA profile for R4 to R6 using true pose\label{fig:aoa_r5_r4}]{%
    \includegraphics[width=0.50\textwidth]{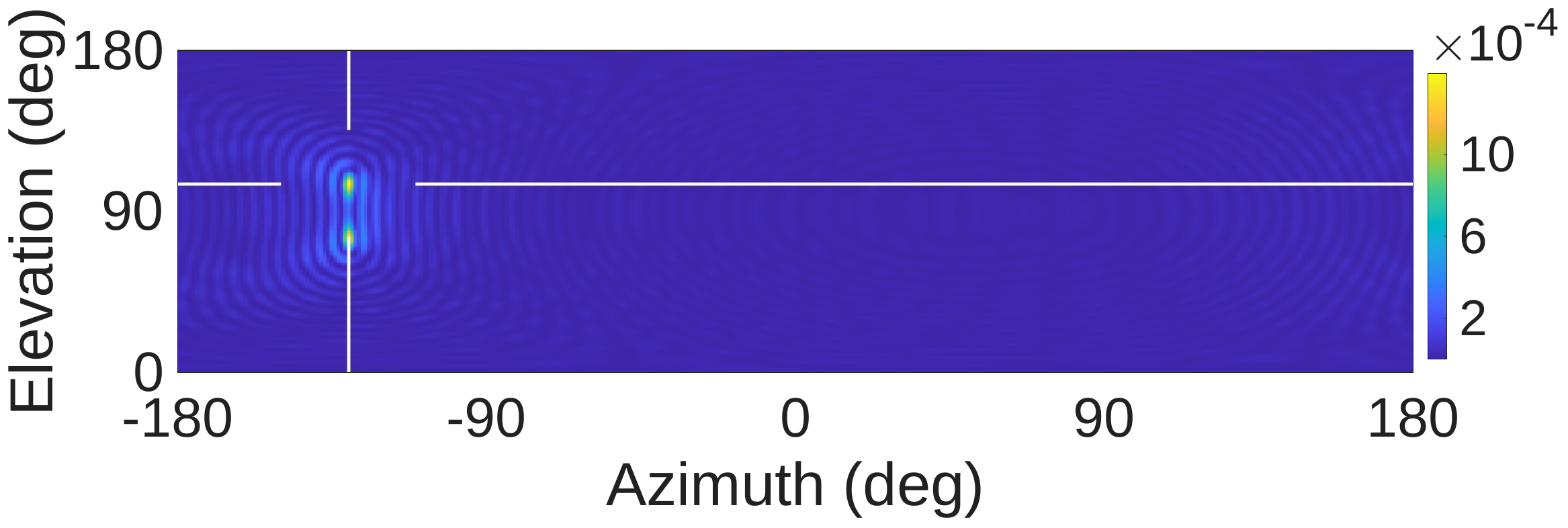}%
}
\caption{AoA profiles for R4 on a 2D circular trajectory, with yellow spots marking AoA peaks and crosshairs indicating true bearing. Figures in (a, c) show the connection to R3 (ground level, high noise), and in (b, d) to R6 (elevated, low noise). Calculations use odometry for (a, b) and true simulated positions for (c, d). R4's purely 2D motion causes elevation ambiguity, resulting in mirrored estimations across the 90$^\circ$ trajectory plane.} 
\label{fig:aoa_example}
\end{figure}

The expanded scenario performs comparably to the baseline using true pose which yields a median position error of $0.04$ m and standard deviation of $0.02$ m. There is no significant difference in AoA error between the expanded and baseline; both have a median around $1.5^\circ$ and standard deviation of $1.5^\circ$. The $AoA_{max}$ estimate using the true pose has an error of $<1^\circ$, which is within estimation resolution, explaining the much higher confidence and lower profile variance in the lower figures of \Cref{fig:aoa_example}.
We see no significant performance difference between elevated and non-elevated nodes in the plots or the data. Despite similar AoA performance, the localization performance significantly improves to $0.37$ m compared with the baseline scenario, due to the additional fourth static beacon.

The AWGN phase noise on its own has little impact when the movement of the robot is captured perfectly as it simply lowers the confidence but does not shift the main peak nor make other peaks appear. We do experience more AoA covariance with lower SNR compared to the higher SNR as can be seen in \Cref{fig:aoa_example}. This partially reproduces the effects of an NLOS scenario in real experiments. However, the amount of profile variance present in the simulation is less severe than in the real NLOS experiments due to the absence of additional peaks caused by multipath signals \cite{Jadhav2022ToolboxRobotics}. 

Furthermore, the planar trajectories by the ground robots in a 3D setup with beacon nodes elevated above or below the mobile nodes result in mirrored AoA estimates in elevation. These ambiguities occur both in our simulated and the real data reported in \cite{Jadhav2022ToolboxRobotics} and reflect known limitations stemming from antenna array geometry \cite{Kumar2014AccurateCost,Jadhav2022ToolboxRobotics}.

\subsection{Generalization Experiment: Random Walk Motion}
\label{subsec:generalization}

To test the framework under unstructured and non-repetitive motion, this scenario replaces circular paths with randomized trajectories. The robots operate in the same 3D environment with a comparable antenna array size and properties as the velocities, signal noise, and simulation setup remain identical apart from the trajectory.
The robot drives linearly for a randomized amount of time, rotates within the range of $[-135, -45] \cup [45, 135]^\circ$, and finishes with a linear section, completing each run spanning $3$ s of simulation time.

\subsubsection*{Generalization Scenario Results}

\begin{figure}[ht]
    \centering
    \includegraphics[width=0.85\linewidth]{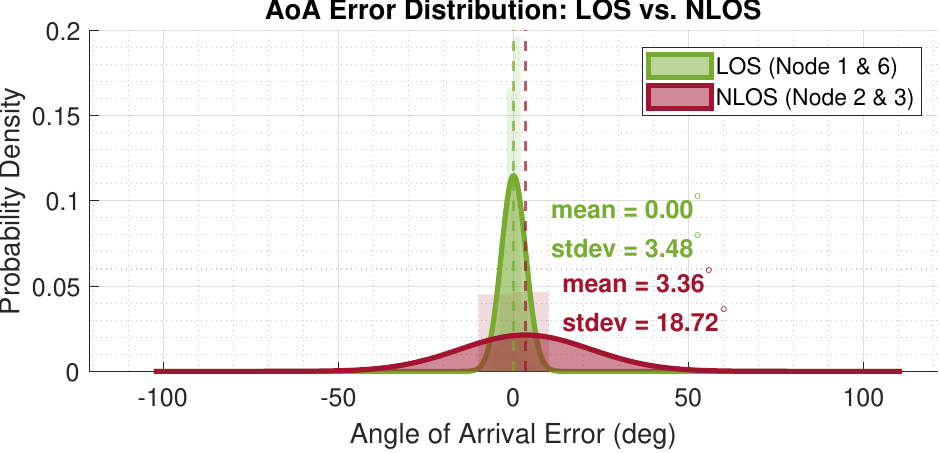}
    \caption{AoA error density for random walk trajectories grouped by level of signal noise; R1 and R6 have an SNR of 30 and are classified to mimic LOS, where R2 and R3 have an SNR of 3, mimicking NLOS scenarios.}
    \label{fig:rwalk-performance}
\end{figure}

The AoA results show a broader spread than circular experiments due to the added trajectory stochasticity (\Cref{fig:rwalk-performance}). The rotational drift in randomized directions better centers the error means at $0^\circ$ and $3.36^\circ$. The standard deviations reach $18.72^\circ$ (high noise) and $3.48^\circ$ (low noise). Performance degradation occurs when noise elevates a grating lobe to AoA$_{max}$. Despite slightly lower than real-world errors ($25.16^\circ$ and $11.43^\circ$ \cite{Jadhav2022ToolboxRobotics}), combining odometry drift with high phase noise produces effects comparable to NLOS multipath interference. Thus, our framework approximates NLOS conditions without the computational overhead of full 3D multipath modeling.

%
%

\section{Discussion \& Conclusion}
\label{sec:conclusion}

Addressing the challenge of relying on deprecated hardware and limited integration with standard robotics tools, we presented a simulation framework integrating MATLAB and Gazebo to evaluate WiFi-CSI-based localization by replicating the WSR toolbox. The framework enables systematic studies of AoA and position estimation under realistic robot motion and free-form trajectories in dynamic environments with nodes at different locations and heights, mimicking ceiling- or wall-mounted access points and multi-floor buildings.

Our environment reproduces key real-world signal properties, including CFO, odometry drift, and additive phase and location noise. Our benchmark's simulated median positioning error of $0.42$ m closely aligns with the $0.50$ m error reported in identical real-world experiments \cite{Jadhav2022ToolboxRobotics}. Furthermore, simulated AoA profiles closely resemble real data in main lobe direction, structure, and patterning. Randomized trajectories mirrored physical AoA error magnitudes, and tunable phase noise successfully approximated both LOS and NLOS conditions, validating its utility for controlled benchmarking. 

The multi-robot results show that increasing static beacons improves localization, while node elevation does not diminish bearing estimation accuracy. 
The mirrored bearing estimates predicted by antenna array geometry further demonstrate the fidelity of our simulation framework. Future work can address multipath-induced sidelobe variations not currently captured by additive noise. 

To ensure scalability and overcome MATLAB's computational overhead, heavy algorithms can be compiled into C/C++ executables or distributed across separate machines via the ROS architecture. 
To conclude, our scalable simulation framework provides a robust, reproducible means to prototype and evaluate WiFi-CSI-based localization algorithms, facilitating research progress and real-world deployments.

\bibliographystyle{spmpsci}
\bibliography{references}
\end{document}